\documentclass[11pt,a4paper,twocolumn]{article}
\usepackage[margin=1in]{geometry}
\usepackage{times}
\usepackage{graphicx}
\usepackage{booktabs}
\usepackage{amsmath}
\usepackage{enumitem}
\usepackage{multirow}
\usepackage{float}
\usepackage{hyperref}
\hypersetup{hidelinks}

\title{Multi-Expert Routing for Multi-Domain Low-Resource OCR: A Manchu Case Study}
\author{
Zhan Chen$^{1,2}$\thanks{\texttt{chenzhan4321@gmail.com}} \quad
Jiqiao Ma$^{3}$\thanks{\texttt{jackiemorin@163.com}} \quad
Chih-wen Kuo$^{4}$\thanks{Corresponding author: \texttt{ziwenkuo@gmail.com}} \\[5pt]
\begin{minipage}{0.9\textwidth}
\centering\small
$^{1}$Institute of Advanced Studies, Beijing Normal University, Zhuhai, China\\
$^{2}$Faculty of Humanities and Social Sciences, Beijing Normal University--Hong Kong Baptist University United International College (UIC), Zhuhai, China\\
$^{3}$Independent Scholar\\
$^{4}$Department of Applied History, National Chiayi University, Chiayi, Taiwan
\end{minipage}
}
\date{July 2026}

\begin{document}
\maketitle

\begin{abstract}
Historical Manchu OCR must accommodate various visually distinct writing styles---like regular script, running script, and the semi-cursive chancery hand used in palace memorials---despite limited labeled data. We study a multi-expert system that reuses checkpoints from an iterative fine-tuning process as domain specialists and uses a lightweight page-level image classifier to dispatch pages by visual style. When the checkpoint pool lacks a suitable specialist, we train an additional expert for that domain.

On three frozen test sets, the routed system matches the selected specialist for each style at two-decimal precision: 0.30\% CER on regular script, 1.57\% on memorials, and 4.83\% on running script. The router achieves 99.3\% page-level domain accuracy and matches the domain-label oracle at the same precision. Two of the three selected specialists were not trained specifically for their final domain; only the running-script expert was trained with that domain as its target. We report the evaluation protocol, router design, and per-page predictions to make the comparison reproducible.
\end{abstract}

\section{Introduction}
Optical character recognition is well developed for many modern, high-resource scripts, but remains difficult for languages represented mainly in historical documents and manuscripts. Existing work in this setting commonly trains one recognizer per language, collection, or hand---from Tesseract \cite{smith2007tesseract} for print to Transkribus \cite{kahle2017transkribus} and kraken \cite{kiessling2019kraken} for historical manuscripts---using the limited annotated data available. This approach is reasonable when the visual material is homogeneous. It becomes less suitable when a single language is represented by several scripts, hands, periods, or media, because the recognizer must then accommodate substantially different image distributions.

Manchu provides a clear example of this problem. It was the administrative language of the Qing empire (1636--1912), and its records are important primary sources for the history of China, Inner Asia, and the early modern world \cite{crossley1993profile,elliott2001manchu}. The surviving material is extensive: the First Historical Archives of China holds on the order of two million documents written wholly or partly in Manchu, while the Qing archive retrieval system of the National Palace Museum in Taipei catalogues 17,056 Manchu-language records, including 2,801 palace memorials. Only a small fraction of this material has been transcribed into machine-readable text. Page-level OCR therefore offers a practical route from digitized images to searchable text, provided that it can accommodate the visual variation within the corpus.

Historical Manchu is also a demanding OCR target (Section~\ref{sec:background} details the script and its material): a vertical, connected script with position-dependent letterforms, degraded archival scans, and material that spans several visually distinct \emph{domains}, from the upright letterforms of woodblock print to the semi-cursive chancery hand of palace memorials to fluent running-script manuscript. Performance on regular-script pages therefore need not transfer to running-hand cursive.

This within-language variation is the central problem considered here and is common to other historical scripts. Pooling all domains into one training set can produce a useful generalist, but it also creates a compromise: in our experiments, a checkpoint that performs well on regular script performs poorly on running script, and the converse also occurs. The limited size of each domain makes it undesirable to train every specialist independently from scratch.

Our starting point is the checkpoint history produced by iterative fine-tuning. The project contains a sequence of versions (v3.0, v4.0, v4.1, and later versions), trained on different data snapshots and augmentation recipes. Their domain scores differ: one checkpoint performs best on regular script, another on memorials, and another on running script. Changes that improve a handwriting domain can therefore reduce performance on a domain that was already well represented. The resulting checkpoints form an observed trade-off in the three-domain error space (Section~\ref{sec:motivation}).

We retain these checkpoints and select complementary specialists from them. A lightweight image-level \emph{domain router} predicts the visual style of each page and dispatches the page to the corresponding specialist. In this finite checkpoint pool, the domain-label oracle combines the lowest observed domain-specific errors, and the learned router matches it at the reported precision. The additional computation is one classifier pass before page transcription; the page recognizers themselves are existing checkpoints.

On the Manchu test sets used here, the routed system reaches sub-percent CER on regular-script pages while also evaluating memorial and running-script material. The broader implication is methodological: in a low-resource, multi-domain project, earlier checkpoints can remain useful when their domain-specific strengths are preserved and selected at inference time. We report the model assignments, router protocol, evaluation procedure, and per-page predictions so that the result can be checked and extended.

Our contributions are:
\begin{enumerate}[noitemsep]
    \item a benchmark-scoped trade-off formulation of multi-domain low-resource OCR, in which checkpoints from a fine-tuning \emph{version stream} are retained as candidate domain specialists;
    \item a lightweight image-level domain router that dispatches pages to experts, with an explicit leakage-safe training protocol, a source-balancing fix for visually adjacent domains, and a domain-label oracle;
    \item empirical evidence on historical Manchu that routing matches the best evaluated domain scores at two-decimal precision (regular 0.30\%, memorial 1.57\%, running 4.83\% CER), which no single checkpoint attains, using two checkpoints not trained specifically for their selected domains and one purpose-trained running-script checkpoint; and
    \item a reproducible evaluation package consisting of the model assignments, router protocol, source-separated test sets, and per-page predictions.
\end{enumerate}

\section{Background}
\label{sec:background}
\paragraph{The Manchu script and its romanization.} Manchu is written in an alphabetic script adapted, by way of Mongolian, from Old Uyghur; it runs vertically from top to bottom, with columns ordered left to right. It is a connected script: letters join into a continuous word-stem, and most letters take distinct initial, medial, and final forms according to position, with further contextual variants. The standard scholarly transcription is the M\"ollendorff romanization, which maps the script to a Latin alphabet with a few diacritics (e.g., \v{s}, \=u, \v{z}). Following common practice and the prior benchmark \cite{chung2025manchu}, our system outputs M\"ollendorff romanization rather than Manchu Unicode: romanization is what most Manchu scholarship reads and searches, and it sidesteps the still-unsettled digital rendering of the script. The recognition target is therefore a Latin character string, and our metric is a character error rate over it.

\paragraph{The written record and its visual styles.} Surviving Manchu materials were produced using different technologies and for different purposes. Manchu palaeography therefore distinguishes several writing styles rather than one uniform hand; the First Historical Archives of China groups much of its material into categories such as memorial-petition, book, proclamation, running-hand, and woodblock styles. For recognition we group our corpus into three visual styles:
\begin{itemize}[noitemsep]
\item \emph{regular script}---the upright, well-separated letterforms of woodblock and movable-type imprints (the archives' woodblock style) and of formal book hands (its book style): reign chronicles, primers, printed collections, and neatly written archival pages;
\item \emph{memorial script}---the controlled semi-cursive chancery hand of palace memorials submitted to the throne (the archives' memorial-petition style), frequently bearing vermilion imperial rescripts;
\item \emph{running script}---the fluent, ligature-heavy semi-cursive of manuscripts and correspondence (the archives' running-hand style).
\end{itemize}
Although these materials use the same language and script, their letterforms and page layouts differ substantially. Our corpus---the Old Manchu Archives, the \texttt{manc.hu} manuscript database, printed collections, and palace memorials---therefore spans all three styles rather than a single clean source.

\section{Related Work}
\paragraph{OCR for ancient and historical scripts.} Recent work on OCR and HTR for historical and low-resource scripts addresses similar data constraints. Examples include cuneiform sign detection with annotated corpora and processing pipelines \cite{ebl2026cuneiform}; new datasets for East Syriac \cite{khamis2024syriac} and Sahidic Coptic \cite{scam2026coptic}; recognition of Ge'ez/Ethiopic \cite{hhd2024ethiopic}; Dead Sea Scroll palaeography \cite{popovic2021deadsea}; and Arabic-script resources using open corpora \cite{openiti_makhzan} and multimodal models \cite{qari2026arabic}.

\paragraph{OCR for Asian and minority scripts.} Related work covers palm-leaf datasets \cite{kesiman2016amadi}, Indic layout parsing \cite{prusty2019indiscapes}, Old Uyghur transcription with vision--language fine-tuning \cite{olduyghur2025}, and routed adapters for Tibetan, Yi, Shui, and Dongba \cite{omniocr2026}. Broader benchmarks continue to show substantial variation in performance across scripts \cite{glotocr2026}.

\paragraph{Methods for the low-resource regime.} Common strategies include transfer learning and parameter-efficient fine-tuning of pretrained backbones, such as TrOCR \cite{li2023trocr}; synthetic data generation; few-shot learning from glyph exemplars \cite{souibgui2022fewshot}; and data augmentation \cite{augsurvey2025}. Most systems are organized around one model per language, collection, or hand \cite{nikolaidou2022survey}. Our setting differs in that one language contains several visual domains, and the training history itself supplies a pool of checkpoints with different domain strengths.

\paragraph{Manchu OCR.} Machine-learning work on Manchu OCR is recent, and existing systems are mostly \emph{word-level}. One line of work segments pages into word images---using projection profiles or stroke growth---and recognizes each word with a CRNN \cite{shi2016end}, a sliding-window CNN \cite{zhang2021manchu}, or related models; ManchuOCR is an open-source example \cite{manchuocr}. More recent work fine-tunes vision--language models on synthetic word images and evaluates them at the word level \cite{chung2025manchu}. Institutional recognizers also exist but are not publicly available; as far as we understand, they likewise use a segment-then-recognize design.

\paragraph{Document image-to-sequence models.} Page-level recognizers avoid an explicit segmentation stage by encoding a page image and emitting the transcription directly. Donut \cite{kim2022donut} and Nougat \cite{blecher2023nougat} are examples of document encoder--decoder models; Pix2Struct \cite{lee2023pix2struct}, Kosmos-2.5 \cite{lv2023kosmos25}, and GOT-OCR2.0 \cite{wei2024got} use related document or screenshot-reading objectives. General-purpose vision--language models \cite{bai2023qwen,liu2024visual} can also transcribe pages, although specialized document models remain useful for high-precision settings. Line- and word-level recognizers, including CRNN/CTC \cite{shi2016end}, TrOCR \cite{li2023trocr}, and kraken \cite{kiessling2019kraken}, require an upstream segmentation stage that page-level models avoid.

\paragraph{Mixtures of experts, ensembles, and reusing checkpoints.} Routing inputs to specialized models is a classical idea \cite{jacobs1991adaptive}, and modern mixture-of-experts systems learn routers and experts jointly \cite{shazeer2017outrageously,fedus2022switch,riquelme2021vmoe,jiang2024mixtral,dai2024deepseekmoe}. Our system operates at the model level: the experts are independently trained checkpoints and the router is an external classifier. Snapshot ensembles retain several checkpoints from one training run \cite{huang2017snapshot}, whereas model soups and stochastic weight averaging merge model parameters \cite{wortsman2022soups,izmailov2018swa}. These methods generally target one data distribution. We instead use checkpoints with different domain scores and select one model per page, so the comparison concerns domain specialization rather than output averaging or weight merging.

\section{Task and Evaluation Protocol}
The task is image-to-text transcription of a complete Manchu page or folio into M\"ollendorff-style Romanization. Input images follow a fixed convention recorded with the data, including image enhancement and optional cropping or denoising. The original orientation and reading order are preserved for evaluation and human review.

For a page with reference string $g$ and prediction $p$, character error rate is
\begin{equation}
    \mathrm{CER}(p,g)=\frac{\mathrm{Lev}(p,g)}{|g|},
\end{equation}
with whitespace normalized consistently and the page evaluated in its original reading order. We report the mean over pages. This is a page-level CER, \emph{not} a word accuracy, and must not be compared directly with word-image benchmarks (Section~\ref{sec:priorwork}).

\paragraph{Frozen, source-separated test sets.} Test reporting uses three domain-specific sets held fixed across the project so that successive versions remain comparable:
\begin{itemize}[noitemsep]
    \item \textbf{regular script}: 125 pages of regular, well-separated letterforms from several sources (woodblock and movable-type collections and the regular hand of the Old Manchu Archives), the principal regular-script test set;
    \item \textbf{memorial}: 10 handwritten palace-memorial pages (a memorial-domain probe);
    \item \textbf{running script}: 13 semi-cursive manuscript pages (a running-script probe).
\end{itemize}
Each set is derived directly from the \texttt{test} split of the master data registry by a single construction script, so it remains synchronized with the frozen split and is never drawn from a stale snapshot. The handwriting sets are small and are treated as domain-shift probes rather than large-sample estimates; enlarging them with additional expert-transcribed held-out pages is ongoing parallel work, kept separate from these frozen sets to preserve comparability.

\section{Domains and the Version Stream}
\label{sec:motivation}
The project maintains a source-aware data registry (about 9{,}054 page/folio records across nine sources as of version v4.1, and still growing) with a permanent train/validation/test split; the three test sets above are frozen subsets and are not used for model training or router checkpoint selection. They are, however, used retrospectively to designate the expert pool after the training runs are complete, as disclosed in Section~\ref{sec:method}. Successive training versions differ in their data snapshot and augmentation recipe. In particular, regular-script sources use mild augmentation, whereas the memorial and running-script sources receive stronger, style-specific augmentation and source-aware oversampling, so that later versions increasingly favor the harder handwriting styles.

The key empirical observation is that \emph{no single version is best on all three test sets}. Table~\ref{tab:versions} reports page-level CER for representative Nougat checkpoints. The strongest regular-script and memorial checkpoints are v3.0 and v3.0\,s2, respectively; neither was trained specifically for the final assignment. The best earlier checkpoint on running script, v4.1\,s2, reaches only 26.68\%. A later v5.0 run trained with running-script data reduces this error to 4.83\%, but its regular-script and memorial scores remain above those of the earlier checkpoints. These results provide the basis for selecting different checkpoints for different domains.

\begin{table}[t]
\centering
\caption{Page-level CER (\%) of representative Nougat checkpoints on the three frozen test sets (regular P125, memorial Z10, running C13). \textbf{Bold} marks each test set's best result (column minimum). No single version wins every column. The upper group contains checkpoints selected after training; the lower group contains purpose-trained v5.0 variants.}
\label{tab:versions}
\resizebox{\columnwidth}{!}{%
\begin{tabular}{lccc}
\toprule
Version & Regular & Memorial & Running \\
\midrule
\multicolumn{4}{l}{\emph{checkpoints selected after training}}\\
v3.0          & \textbf{0.30} & 3.40 & 47.09 \\
v3.0 (seed 2) & 0.30 & \textbf{1.57} & 51.59 \\
v4.1 (seed 2) & 0.45 & 3.65 & 26.68 \\
\midrule
\multicolumn{4}{l}{\emph{deliberate (purpose-trained v5.0)}}\\
v5.0 (mem.-tuned)  & 0.41 & 1.60 & 6.86 \\
v5.0 (curs.-tuned) & 0.79 & 3.29 & \textbf{4.83} \\
\bottomrule
\end{tabular}%
}
\end{table}

\begin{figure}[t]
\centering
\includegraphics[width=\columnwidth]{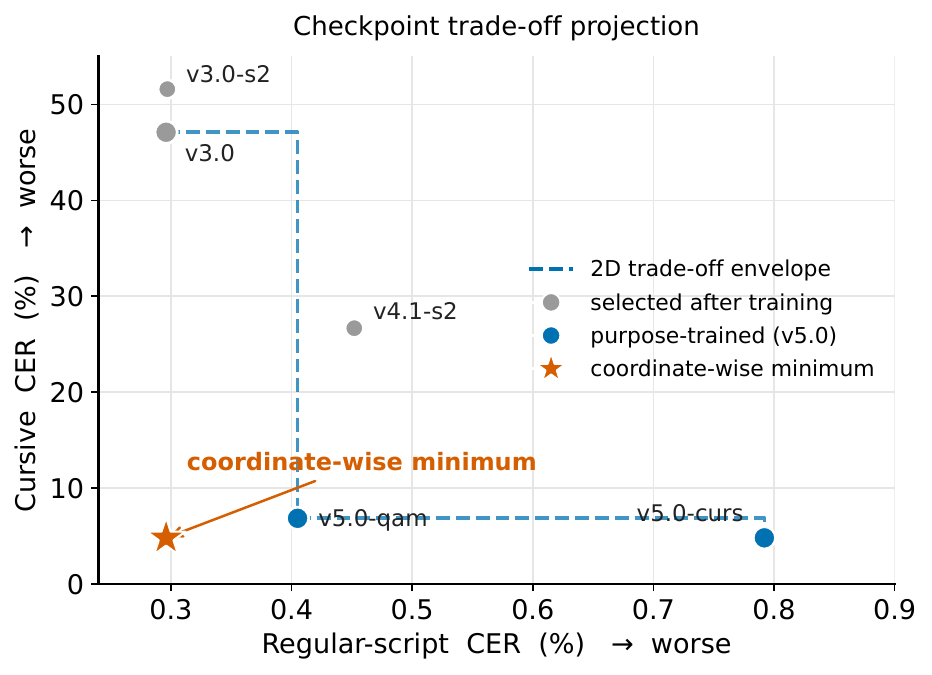}
\caption{A two-dimensional projection of the checkpoint trade-off: regular-script CER versus running-script CER (from Table~\ref{tab:versions}). Grey points are checkpoints selected after training; blue points are purpose-trained v5.0 variants. The red star marks the coordinate-wise minimum on the two displayed test sets. The plot does not characterize the full three-set error space.}
\label{fig:pareto}
\end{figure}

\paragraph{A benchmark-scoped trade-off view.} Figure~\ref{fig:pareto} projects errors on two of the three test sets, with lower CER indicating better performance. It visualizes the observed checkpoint trade-off rather than establishing non-dominance in the full three-dimensional space. The coordinate-wise minima in this pool are $(0.30, 1.57, 4.83)$ for regular, memorial, and running script, respectively. No single checkpoint attains all three values. The domain-label oracle combines the corresponding specialists, and the learned router matches it at two-decimal precision. The result is specific to the checkpoints and test sets used here and should be reassessed as the data and model pool expand.

\section{Method: Version Experts and a Domain Router}
\label{sec:method}

\begin{figure*}[t]
\centering
\includegraphics[width=0.96\textwidth]{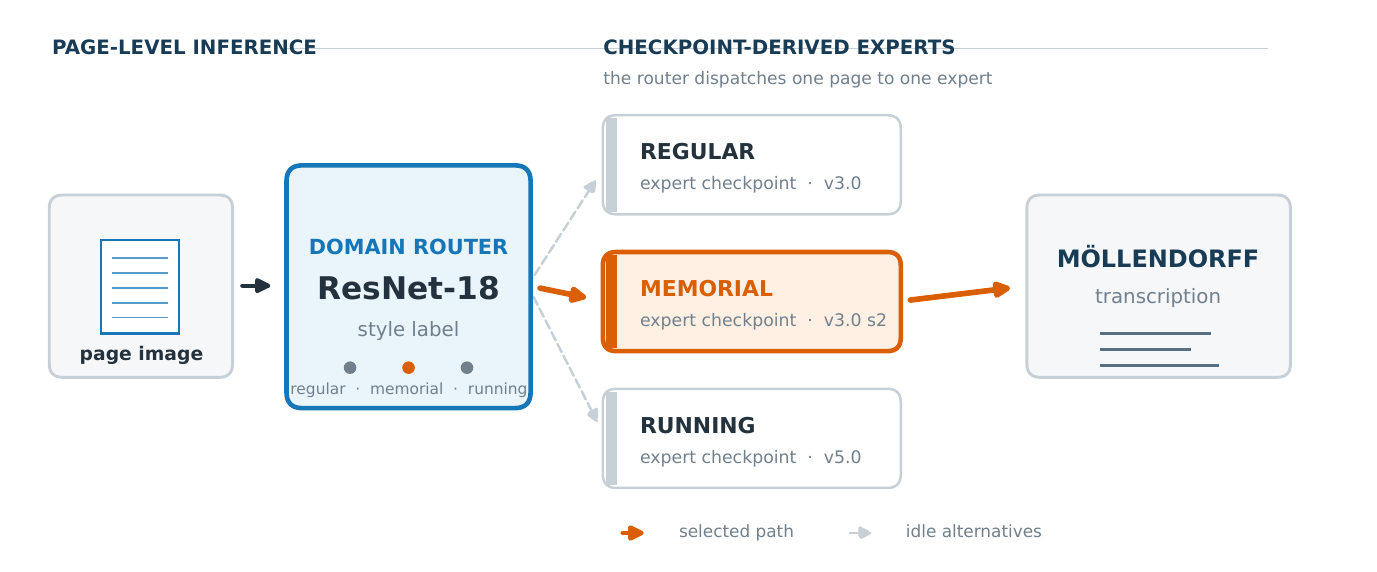}
\caption{The routed system. A lightweight domain router (a small image classifier) reads each page and predicts its visual style; the page is then transcribed by the corresponding domain expert---a Nougat checkpoint drawn from the project's own version stream. The solid orange path shows a memorial page routed to the memorial expert (v3.0\,s2); the dashed gray branches are alternative experts left idle for that page.}
\label{fig:system}
\end{figure*}

\paragraph{Problem statement.} Figure~\ref{fig:system} sketches the routed system. Let a page image $x$ belong to one of $K$ visual domains, and let $\{e_1,\dots,e_M\}$ be a pool of transcription experts, each a complete page-to-text model. A \emph{router} $r$ maps $x$ to an expert index, and the system transcribes $x$ as $e_{r(x)}(x)$. Writing $c(e,x)$ for the CER of expert $e$ on page $x$, we distinguish two reference policies. The \emph{domain-label oracle} uses the true domain label $y(x)$ and dispatches to its designated expert $e_{y(x)}$; this is the oracle reported in our experiments. A stronger \emph{per-page oracle} would choose $\arg\min_m c(e_m,x)$ and is only a theoretical upper bound, not an experimental baseline here. With one designated specialist per domain, the learned router equals the domain-label oracle when its domain prediction is correct, but it need not equal the per-page oracle. The system therefore reduces to domain prediction followed by expert selection. Since the experts are existing checkpoints, the added model is only the page-level classifier.

\paragraph{Expert pool.} All experts share one architecture---a Nougat document encoder--decoder \cite{blecher2023nougat} adapted to Manchu Romanization---and differ in their training version. The appendix reports a common-recipe comparison of three candidate backbones; Nougat performed most consistently across the three styles. From this pool we designate three domain experts, each the best available checkpoint on its domain in Table~\ref{tab:versions}: v3.0 for regular script, v3.0\,s2 for memorials, and v5.0 for running script. The first two checkpoints were selected after training rather than trained for those final assignments; the v5.0 running-script checkpoint was trained specifically for that domain. Because the expert pool was designated retrospectively from the completed test-set results, the routed number is a frozen-benchmark comparison rather than an independently held-out estimate of expert-pool selection.

\paragraph{Domain router.} The router is a small residual image classifier \cite{he2016deep} (ResNet-18, ImageNet-initialized, $224\times224$ input) that maps a page image to a domain label (regular / running / memorial). At inference the page is transcribed by the expert corresponding to the predicted label. Because reading a low-resolution page is enough to judge visual \emph{style}, the router is small and fast; it adds a single classifier forward pass per page.

\paragraph{Source-balanced training.} The router's supervision is imbalanced: the training split contains roughly $8000$ regular-script, $153$ running-script, and $50$ memorial pages. Inverse-frequency class weighting and weighted sampling address the class imbalance. A second imbalance occurs within the regular-script class, which is dominated by one archival source. A naive router underfits the visually distinct manc.hu woodblock print and confuses it with running-hand pages from the same collection. We therefore assign equal sampling mass to regular-script sources. In the ablation, this change reduces the resulting confusion between visually adjacent sources.

\paragraph{Leakage-safe training.} The router is trained and selected only on training-split pages; the frozen test sets (regular, memorial, running) are not used for router fitting or router checkpoint selection. Domain labels are derived from source metadata (running-script source $\rightarrow$ running; memorial source $\rightarrow$ memorial; all others $\rightarrow$ regular). Because project policy removes the memorial source from the global validation pool, we reserve a small held-out slice of \emph{training} memorial pages purely to monitor memorial recall during router checkpoint selection; these pages are never part of the memorial test set. Separately, the expert pool itself was designated retrospectively from the completed frozen-test-set results, so this experiment does not claim a fully blind estimate of model-selection performance.

\paragraph{Domain-label oracle.} To separate router error from expert quality, we report a domain-label oracle that uses each page's true domain label and the fixed domain-to-expert mapping. The gap between the learned router and this oracle measures the cost of domain-routing mistakes; it is distinct from the stronger per-page oracle defined above.

\section{Experiments}
\label{sec:experiments}
We evaluate on the three frozen test sets. For every page we run each expert, record its CER, and then report, for each set, every single expert applied to the full set, the domain-label oracle, and the learned router.

\begin{figure}[t]
\centering
\includegraphics[width=0.82\columnwidth]{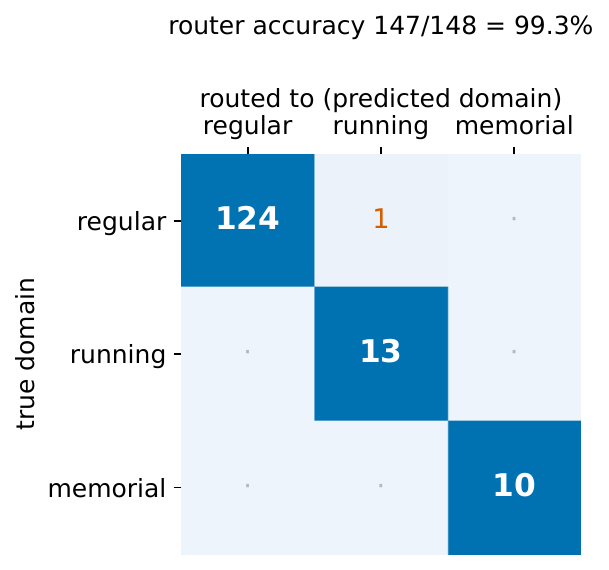}
\caption{Domain-router confusion matrix on the frozen test sets (rows: true domain; columns: expert routed to). The learned router places 147 of 148 pages correctly (99.3\%); the single error is one atypical manc.hu regular-script page sent to the running-script expert. Its regular-script CER is 0.299\%, versus 0.296\% for the domain-label oracle, which rounds to 0.30\% for both.}
\label{fig:confusion}
\end{figure}

\paragraph{Routing matches the domain-label oracle at reported precision.} With three experts---regular script (v3.0), running script (v5.0), and memorial (v3.0\,s2)---the router dispatches pages to the corresponding specialist. It matches the domain-label oracle to two decimal places on all three test sets: 0.30\% on regular script, 1.57\% on memorials, and 4.83\% on running script (Table~\ref{tab:router3}). Its domain accuracy is 99.3\% (147/148; Figure~\ref{fig:confusion}). The single-expert results show why selection matters: the regular-script expert reaches 47.1\% CER on running script, whereas the running-script expert reaches 0.79\% on regular script.

\begin{table}[t]
\centering
\caption{Three-expert routing (regular v3.0 + running v5.0 + memorial v3.0\,s2), page-level CER (\%) on the frozen test sets (regular P125, memorial Z10, running C13). The router matches the domain-label oracle to two decimal places on all three sets. Domain accuracy is 99.3\% (147/148). Off-diagonal columns show the cost of sending a page to the wrong expert.}
\label{tab:router3}
\resizebox{\columnwidth}{!}{%
\begin{tabular}{lcccccc}
\toprule
Test set & \shortstack{reg\\only} & \shortstack{run\\only} & \shortstack{mem\\only} & oracle & \textbf{router} & mis \\
\midrule
regular   & 0.30  & 0.79 & 0.30  & 0.30 & \textbf{0.30} & 1/125 \\
memorial & 3.40  & 3.29 & 1.57  & 1.57 & \textbf{1.57} & 0/10  \\
running   & 47.09 & 4.83 & 51.59 & 4.83 & \textbf{4.83} & 0/13  \\
\bottomrule
\end{tabular}%
}
\end{table}

\paragraph{Effect of domain-specific training.} The comparison separates the contribution of the checkpoint history from that of the purpose-trained expert. The best earlier checkpoint scores 26.68\% on running script; the running-script v5.0 checkpoint reduces this to 4.83\%. The same v5.0 campaign does not replace the selected checkpoints on regular script or memorials: its best scores on those test sets are 0.41\% and 1.60\%, compared with 0.30\% and 1.57\%. The selected pool therefore combines checkpoints from different stages of training rather than using a single final version.

\begin{figure}[t]
\centering
\includegraphics[width=\columnwidth]{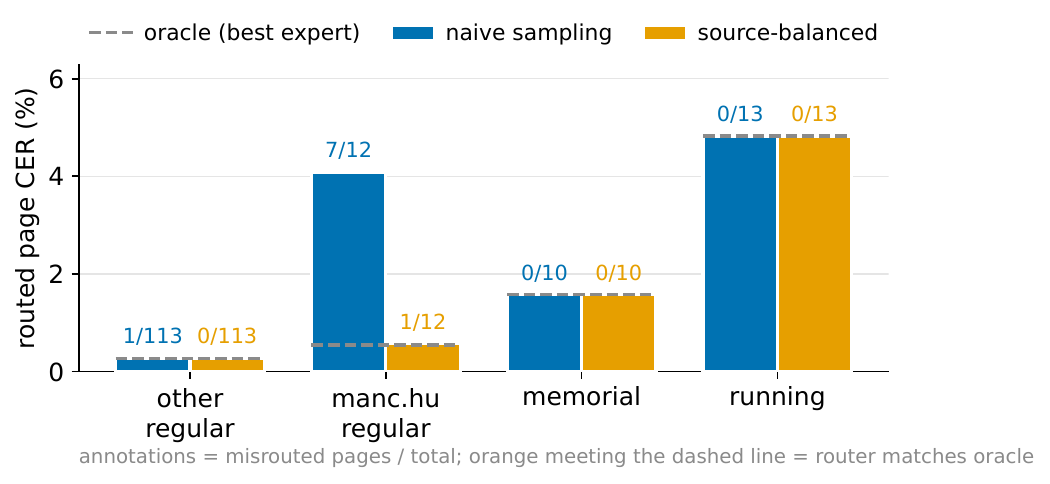}
\caption{Source-balanced router training (ablation) on the frozen test sets. Within the regular-script class, a naive router underfits the minority manc.hu woodblock print and misroutes most of it to the running-script expert (blue, 7/12), inflating its routed CER to 4.1\% against a domain-label oracle of 0.5\%; balancing the regular-script sources (orange) recovers domain-label-oracle-level routing (1/12, 0.6\%). The dominant archival regular hand and the other styles are unaffected.}
\label{fig:ablation}
\end{figure}

\paragraph{Router errors and source balancing.} One page is misrouted: an atypical manc.hu regular-script page is sent to the running-script expert. The effect on the regular-script test set is small (0.299\% routed CER versus 0.296\% for the domain-label oracle; both round to 0.30\%). The ablation shows a larger failure mode when regular-script sources are not balanced: the router confuses manc.hu woodblock pages with running-hand pages from the same collection (Figure~\ref{fig:ablation}). Equalizing the sampling mass of regular-script sources removes most of this confusion.

\paragraph{Why a shared Nougat backbone.} All experts share the Nougat encoder--decoder. Under a common recipe, a from-scratch ResNet-50 recognizer is comparable to Nougat on regular script (0.48\% versus 0.56\%) but performs poorly on handwriting (5.04\% on memorials and 18.67\% on running script). Donut (Swin\,+\,mBART) is weaker on regular script (1.23\%) but reaches 5.89\% on running script (Appendix~\ref{app:arch}). These results support using Nougat for the pooled experts.

\section{Discussion}
The benefit of routing is largest when domain-specific experts differ substantially. On the regular-script test set, routing to v3.0 gives the same rounded CER as applying that expert to every page. On running script, the purpose-trained expert reduces CER without changing the selected regular-script and memorial checkpoints. This separation allows a new domain to be added as an expert and a router class without replacing the existing recognizers.

Within this benchmark, a new specialist can lower the routed error when it improves a domain not already covered by the selected pool. Future work includes stronger running-script models, specialists for seals or marginalia, and larger handwriting test sets so that these comparisons can be estimated with greater precision.

\section{Relation to Prior Manchu OCR Results}
\label{sec:priorwork}
The most relevant prior system fine-tunes a vision--language model on synthetic word images and evaluates at the word level on regular script \cite{chung2025manchu}. An open-source predecessor, ManchuOCR \cite{manchuocr}---a community code release without an accompanying paper---likewise operates at the word level, segmenting a page into word images and recognizing each. Because our system instead transcribes complete pages across three visual styles, neither is directly comparable to it. (At the word level on regular script, our page-to-page system reaches 98.86\% word accuracy, for reference.) These lines of work are complementary; a shared public word-and-page benchmark would allow comparison on common data.

\section{Limitations and Release Status}
The handwriting test sets are small (the memorial and running-script sets have 10 and 13 pages), so they are domain-shift probes rather than large-sample estimates; enlarging them with additional expert-transcribed held-out pages is ongoing. The experts come from successive data snapshots rather than a single controlled recipe sweep, so version differences confound data and augmentation. The router uses a coarse three-way domain label; finer style distinctions may require more classes and more monitoring data, especially for domains removed from the validation pool. Finally, this is a proof of concept with a small number of experts; scaling the pool and the router to more domains is future work.

Further progress depends on collaboration with fluent Manchu readers. In particular, expert review of difficult pages, correction of transcription errors, and expansion of running-script annotations would help improve recognition accuracy where the current system remains weakest.

We also welcome collaborators who would like to join our team. Our broader work focuses on natural-language processing for low-resource languages, not only Manchu, with the aim of helping preserve cultural heritage. Previous outputs have addressed Syriac, Latin, Greek, and Manchu. We also welcome contact from researchers and cultural-heritage projects working with other low-resource languages and related NLP needs.

The release plan has four stages: (1) freeze the split and preprocessing manifest; (2) publish the router, evaluation scripts, and per-page prediction files; (3) publish expert weights subject to the underlying data and model licenses; and (4) add a public, independently adjudicated handwriting benchmark. This arXiv version reports the current state and does not imply that the later stages are complete. A public trial interface for the current page-level OCR is available at \url{http://124.223.33.6/manchu-ocr/}. Translation is being developed as a separate extension and is not evaluated in this paper.

As a next-version extension, we are exploring a three-checkpoint ensemble within each visual domain: v3.0, res50-print, and Donut for regular script; qam-s42, res50-bal, and Donut for memorials; and curs-s43, res50-print, and Donut for running script. The current design uses anchored voting for regular script and memorials and ROVER decoding for running script, with provisional target CERs of 0.28\%, 1.39\%, and 3.81\%, respectively. These targets are not part of the present benchmark results; they will be formally evaluated under the frozen P125/Z10/C13 protocol in a future version.

A companion Manchu-language parser is available at \url{http://124.223.33.6/manchu-parser/}. It currently supports parsing, translation, and word-form analysis. The first four integrated dictionaries come from the Tohoku University collection at \url{http://hkuri.cneas.tohoku.ac.jp/project1/manchu/list?groupId=11}, with additional data from \url{https://manc.hu/}. The parser is a separate language resource and is not part of the OCR benchmark reported here.

\section{Conclusion}
We presented a multi-expert routing approach for multi-domain low-resource OCR. The system reuses checkpoints from an iterative fine-tuning process as domain experts and selects among them with a page-level visual classifier. On the frozen Manchu test sets, it matches the best evaluated domain scores at two-decimal precision: 0.30\% on regular script, 1.57\% on memorials, and 4.83\% on running script. The result supports retaining complementary checkpoints and adding a specialist when a domain remains underrepresented. The paper specifies the evaluation protocol and routing analysis; the release of models and data remains subject to the applicable licenses.

\bibliographystyle{plain}
\sloppy
\bibliography{references}

\clearpage
\onecolumn
\appendix

\section{Architecture comparison}
\label{app:arch}
Table~\ref{tab:arch} compares three page-level backbones trained with the same source-balanced (v5.0) recipe on the frozen test sets. Nougat is the most balanced of the three architectures: the from-scratch ResNet-50 recognizer is strong on regular script but performs poorly on handwriting, while Donut is weaker on regular script. This comparison motivates the use of Nougat for all experts in the routed pool.

\begin{table}[!ht]
\centering
\caption{Page-level CER (\%) of three backbones under a common source-balanced recipe (regular P125, memorial Z10, running C13). Nougat balances all three styles; ResNet-50 is a regular-script specialist; Donut trails on regular script.}
\vspace{6pt}
\label{tab:arch}
\begin{tabular}{l@{\hspace{2.2em}}c@{\hspace{2em}}c@{\hspace{2em}}c}
\toprule
Backbone & Regular & Memorial & Running \\
\midrule
Nougat (enc--dec)        & 0.56 & \textbf{2.33} & 5.92 \\
ResNet-50 (from scratch) & \textbf{0.48} & 5.04 & 18.67 \\
Donut (Swin\,+\,mBART)   & 1.23 & 3.75 & \textbf{5.89} \\
\bottomrule
\end{tabular}
\end{table}

\clearpage
\section{Qualitative examples}
\label{app:samples}
\begin{figure}[H]
\centering
\includegraphics[width=0.86\textwidth]{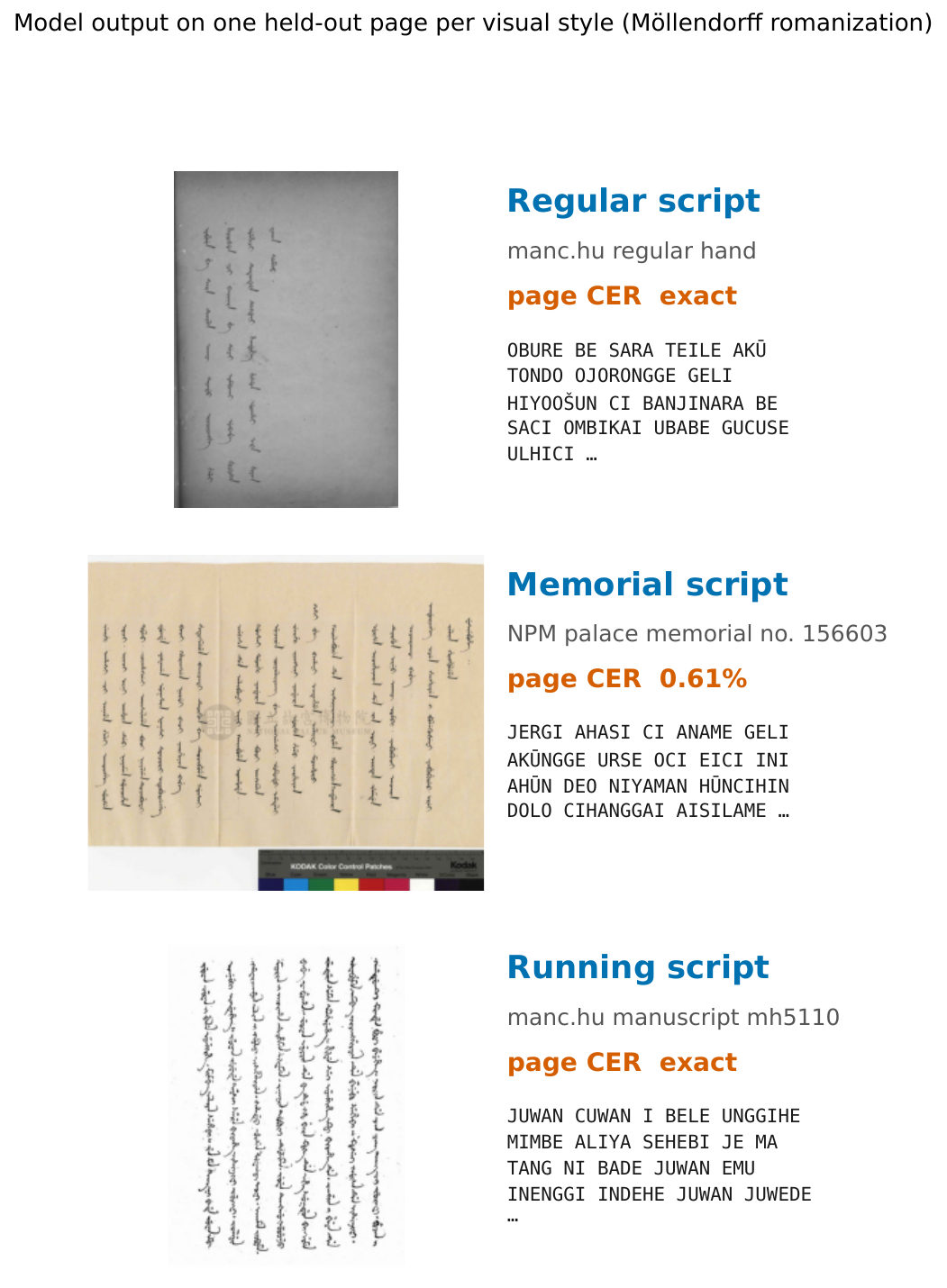}
\caption{Model output on one held-out page per visual style (regular, memorial, running), with the page-level CER and the opening of the predicted transcription. Live showcase: \url{http://124.223.33.6/manchu-ocr/}.}
\label{fig:samples}
\end{figure}

Figure~\ref{fig:samples} shows the routed system's output on one held-out page from each of the three visual styles, in M\"ollendorff romanization. The regular-script and running-script pages are transcribed exactly; the memorial page carries a small residual error rate. These pages are drawn from the frozen test sets and were never seen in training.

\clearpage
\section{Parser showcase}
\label{app:parser}
In addition to the OCR interface, the project provides a public Manchu-language assistance parser at \url{http://124.223.33.6/manchu-parser/}. The tool supports parsing, translation, and full word-form analysis. Its first four integrated Manchu dictionaries come from the Tohoku University collection at \url{http://hkuri.cneas.tohoku.ac.jp/project1/manchu/list?groupId=11}, with additional data from \url{https://manc.hu/}. The example below illustrates the current interface: dictionary glosses are retained in their original Chinese/Japanese form, while the English translation and retrieval-augmented morphological analysis remain experimental.

\begin{figure}[!ht]
\centering
\includegraphics[width=0.98\textwidth]{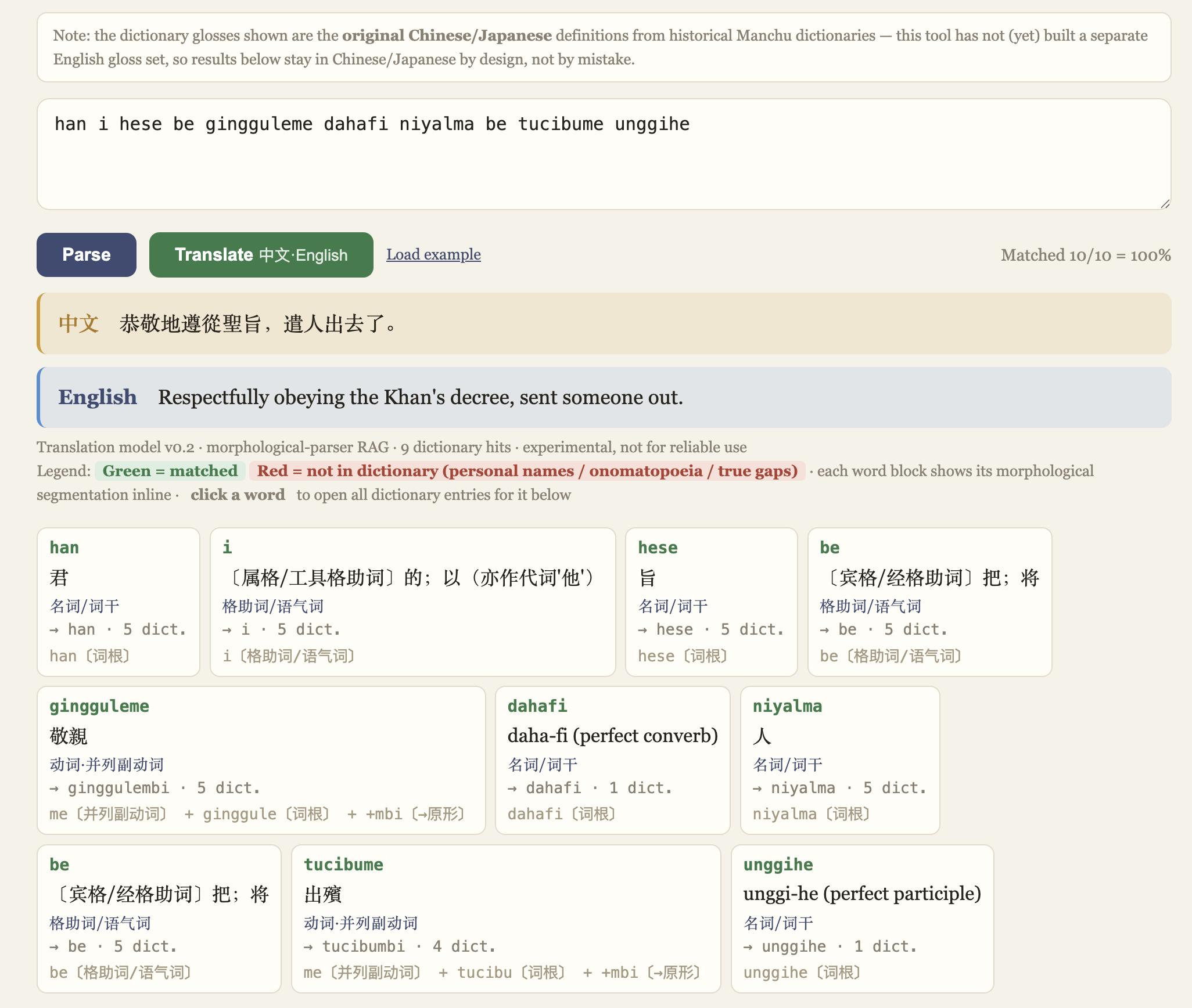}
\caption{Example output from the public Manchu-language assistance parser. The interface shows parsing, Chinese translation, an experimental English translation, dictionary hits, and inline morphological segmentation for one sentence. The displayed ``10/10'' match is an interface example, not an independent accuracy evaluation.}
\label{fig:parser-showcase}
\end{figure}

\end{document}